\documentclass{article}

\usepackage{arxiv}

\usepackage[utf8]{inputenc} 
\usepackage[T1]{fontenc}    
\usepackage{hyperref}       
\usepackage{url}            
\usepackage{booktabs}       
\usepackage{amsfonts}       
\usepackage{nicefrac}       
\usepackage{microtype}      
\usepackage{lipsum}
\usepackage{soul}
\usepackage{amsmath}
\usepackage{url}
\usepackage{multirow}
\usepackage{amssymb}
\usepackage{subfig}
\usepackage{graphicx}

\newcommand{\argmin}{\mathop{\rm argmin}\limits}
\newcommand{\norm}[1]{\left\lVert#1\right\rVert}

\title{\Large City2City: Translating Place Representations across Cities}

\author{
  Takahiro Yabe \\
  Lyles School of Civil Engineering\\
  Purdue University, USA\\
  \texttt{tyabe@purdue.edu} \\
   \And
 Kota Tsubouchi \\
  Yahoo Japan Corporation\\
  Tokyo, Japan\\
  \texttt{ktsubouc@yahoo-corp.edu} \\
   \And
 Toru Shimizu \\
  Yahoo Japan Corporation\\
  Tokyo, Japan\\
  \texttt{tshimiz@yahoo-corp.edu} \\   
  \And
 Yoshihide Sekimoto \\
 Institute of Industrial Science\\
  University of Tokyo\\
  Tokyo, Japan\\
  \texttt{sekimoto@iis.u-tokyo.ac.jp} \\
  \And 
  Satish V. Ukkusuri\\
  Lyles School of Civil Engineering\\
  Purdue University\\
  Indiana, USA \\
  \texttt{sukkusur@purdue.edu} \\
}

\begin{document}
\maketitle

\begin{abstract}
Large mobility datasets collected from various sources have allowed us to observe, analyze, predict and solve a wide range of important urban challenges.
In particular, studies have generated place representations (or embeddings) from mobility patterns in a similar manner to word embeddings to better understand the functionality of different places within a city. 
However, studies have been limited to generating such representations of cities in an individual manner and has lacked an inter-city perspective, which has made it difficult to transfer the insights gained from the place representations across different cities. 
In this study, we attempt to bridge this research gap by treating \textit{cities} and \textit{languages} analogously.
We apply methods developed for unsupervised machine language translation tasks to translate place representations across different cities. 
Real world mobility data collected from mobile phone users in 2 cities in Japan are used to test our place representation translation methods. 
Translated place representations are validated using landuse data, and results show that our methods were able to accurately translate place representations from one city to another.
\end{abstract}


\section*{Note:}
\textbf{This work was accepted as a 4-page short paper in the ACM SIGSPATIAL 2019 Conference. This is the extended (originally submitted) version.  \url{https://dl.acm.org/citation.cfm?id=3359063}}

{\small
\begin{verbatim}
@inproceedings{Yabe:2019:CTP:3347146.3359063,
 author = {Yabe, Takahiro and Tsubouchi, Kota and Shimizu, Toru and Sekimoto, Yoshihide 
           and Ukkusuri, Satish V.},
 title = {City2City: Translating Place Representations Across Cities},
 booktitle = {Proceedings of the 27th ACM SIGSPATIAL International Conference on 
              Advances in Geographic Information Systems},
 series = {SIGSPATIAL '19},
 year = {2019},
 isbn = {978-1-4503-6909-1},
 location = {Chicago, IL, USA},
 pages = {412--415},
 numpages = {4},
 url = {http://doi.acm.org/10.1145/3347146.3359063},
 doi = {10.1145/3347146.3359063},
 acmid = {3359063},
 publisher = {ACM},
 address = {New York, NY, USA},
 keywords = {human mobility, machine translation, mobile phone data, place representations, 
             urban functions},
} 
\end{verbatim}
}

\section{Introduction}
Large mobility datasets collected from mobile phones, social media, and various sensors have allowed us to observe the dynamics of cities at an unprecedented spatio-temporal resolution \cite{batty2012smart,ratti2006mobile}.
Data-driven methods have revolutionized the way we tackle various urban challenges  \cite{zheng2014urban} such as pollution \cite{zheng2013u}, traffic congestion \cite{iqbal2014development}, and disaster management \cite{lu2012predictability}. 

In particular, recent studies have made significant progress on understanding the semantics of places from large mobility data \cite{feng2017poi2vec,liu2016exploring,yan2017itdl,zhao2017geo,wang2017region}. 
Many of these studies produce place representations (or embeddings) in a similar manner to generating word representations in the natural language processing field (e.g. \texttt{word2vec} \cite{mikolov2013distributed}), by treating words and places, sentences and trip sequences as analogies. 
These place representations have been shown to successfully characterize the functionality of places within cities, and have been applied in various tasks related to urban planning, such as identifying spatial clusters with respect to functionality \cite{yao2018representing}, choosing sites for opening new stores \cite{xu2016}, and predicting where users will go to in future timesteps \cite{chang2018content}. 

However, such studies have been limited to understanding the place representations of cities in an individual manner, and has lacked an inter-city perspective. 
Because the representations of different cities were not generated in a common vector space, it has been difficult to transfer insights based on place representations from one city to another, let alone transferring various phenomena (e.g. evacuation after disasters) across cities. 
If we could map place representations learned in one city to another, in the same way we translate words to words across different languages, we would be able to utilize knowledge accumulated in other cities to perform better analyses and predictions.   
For example, we may be able to predict locations that could become evacuation shelters in future disasters in city $\psi$, by translating the representations of places that became evacuation shelters in a past disaster in city $\phi$ to city $\psi$.

\begin{figure*}[!t]
  \centering
  \includegraphics[width=0.75\textwidth]{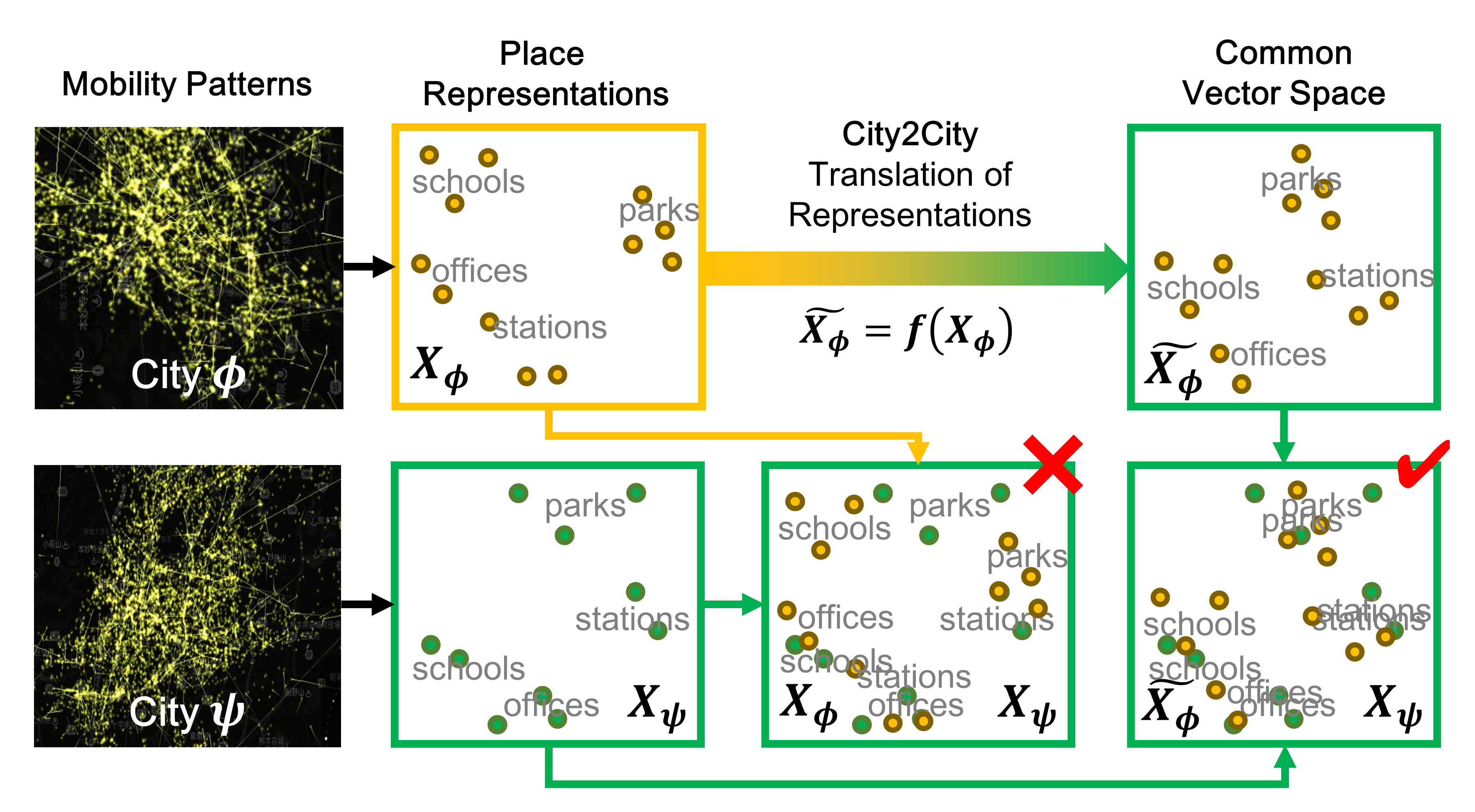}
  \caption{
  Illustration of our problem setting.
  Given representations of places in 2 cities ($\phi,\psi$) generated from the observed mobility patterns, our problem is to translate the place representations of city $\phi$ to the vector space of city $\psi$, so that similar places from the two cities become mapped closely in the common vector space, as shown in the bottom right panel. 
  }
  \label{problem}
\end{figure*}

In this study, we attempt to bridge these gaps by treating \textit{cities} and \textit{languages} analogously, which extends the analogies made by previous studies (``places and words'' and ``trip sequences and sentences'').
More specifically, our goal is to develop methods that can map places from different cities with similar meanings closely on a common vector space via an operation analogous to \textit{translation}. 
We propose models that extend the methods developed in the natural language processing field for unsupervised machine language translation tasks \cite{zhang2017adversarial,conneau2017word,artetxe2018robust}. 
Figure \ref{problem} shows an illustration of our problem setting and approach. 
Given representations of places $X_{\phi}$, $X_{\psi}$ in cities $\phi$ and $\psi$, directly overlaying $X_{\phi}$ on $X_{\psi}$ would be uninformative, since the vector spaces are not aligned with eachother. 
Using methods to translate representations, we obtain $\tilde{X}_{\phi} = f(X_{\phi})$ which is aligned to the space of city $\psi$, allowing us to compare representations of places in different cities for further analyses and predictions. 

The model performances are tested using real world data collected from mobile phones in 2 cities in Japan, and are validated using landuse data.  
Results show that our methods are able to accurately translate place representations from one city to another.

The main contributions of this paper are as follows:
\begin{itemize}
    \item We propose and test methods to translate place representations across cities, which can map places from different cities with similar functions closely together.
    \item We verify that our method can successfully translate place representations, using real mobility world data from 2 cities. 
    \item We make the translated place representations publicly available for researchers and practitioners. 
\end{itemize}


\section{Preliminaries}
\subsubsection*{Definition 1 (\textbf{Human Mobility Patterns})} 
Sequences of users' staypoint locations with timestamps are extracted from mobility data using methods explained in Section 3.1.
The usual human mobility patterns of a city $c$ is the set of all staypoint sequences of individuals whose home location belongs to city $c$.

\subsubsection*{Definition 2 (\textbf{Place Representations})} 
A city $c$ is divided into disjoint cells by grid sizes of $r$ meters. 
We will call each cell as a place $i$, and denote its representation as $\textbf{x}_i^c$, which is a $d$-dimensional vector.
Place representations $\textbf{x}_i^c$ are learned from the human mobility patterns observed in city $c$, using methods explained in Section 3.2. 
Representations of all places are stacked as a $(d\times n_c)$ matrix $\textbf{X}_c$, where $n_c$ is the number of places in city $c$.

\subsubsection*{Problem Definition (\textbf{Translation of Place Representations})} 
Place representations $\textbf{X}_{c}$ are learned for each city $c$ from the observed mobility patterns. 
Thus, for different cities, the vector spaces are not shared. 
Translating place representations from city $\phi$ to city $\psi$ is equivalent to finding a mapping function $f$ that aligns the two vector spaces, i.e., $\textbf{X}_{\psi} \approx \tilde{\textbf{X}_{\phi}}= f(\textbf{X}_{\phi})$. 
Methods used to translate place representations are explained in Section 3.3.



\section{Methodology}

\subsection{Extracting Human Mobility Patterns}
We first extract human mobility pattern datasets for each city, from the location data observed from mobile phones. 
Each observation of the location data contains the user ID, timestamp, longitude and latitude.
More details of the mobile phone data that we use in this study are explained in Section 4.1.1. 
Our goal is to extract users' sequences of staypoint locations from the observations.
We achieve this by setting two threshold parameters; one spatial threshold and one temporal threshold. 
To cope with noisy location observations (e.g. spatial errors in GPS data), we perform mean shift clustering to estimate the true location for each observation, as described in previous studies (e.g. \cite{ashbrook2003using,kanasugi2013spatiotemporal}).  
For each user, we read their location data in time order, and search for locations where the user has stayed within the distance defined by the spatial threshold parameter for a duration longer than the time defined by the temporal threshold parameter. 
We use 1000 meters as the spatial threshold, and 30 minutes as the temporal threshold in this study. 
As a result, we are able to obtain sequences of staypoint locations for each user, which will be used to generate place representations using methods explained in the following section.

\subsection{Generating Place Representations}
To obtain the representations of places in a city, we solve a self-supervised task in which an LSTM RNN model is trained to predict the next staypoint of a user using mobility data, which is analogous to language models which are trained to predict the next word in a sentence.
After training an LSTM RNN model using staypoint sequences of a city $c$, we extract and stack the embedding layer's parameters of the size $n_c \times d$, and define it as the matrix of place representations $\textbf{X}_c$.
We refer to this place representation learning architecture as ``\textbf{\texttt{MobLSTM}}'' in the following sections.
Specific model hyperparameter settings are explained in Section 4.2.1.

\subsection{Translating Place Representations}
Three approaches for translating place representations across cities are tested. 
The first approach is to jointly learn the place representations of places in both cities using a common \texttt{MobLSTM} architecture (Section 3.3.1). 
The second and third approaches learn place representations using \texttt{MobLSTM} separately for different cities, and then attempt to align them using an optimization method (Section 3.3.2), or adversarial training (Section 3.3.3). 

\subsubsection{Joint Learning Approach}
\label{sec:joint_learning}
In the first approach, we apply the \texttt{MobLSTM} model to the two cities together on the self-supervised next staypoint prediction task.
We merge the mobility datasets of two cities into one, train the model over the merged data, and use the transposed embedding layer matrix of the size $d \times (n_{\phi} + n_{\psi})$ as the representation matrix.
The rationale behind this approach is that, representations of places with similar functions will be visited in a similar manner (e.g. time of day, day of week, after and before certain places) regardless of the city the places belong to.
To let the model treat places of cities $\phi$ and $\psi$ as equally as possible, we mask the candidates of $\psi$ at the output when the next staypoint belongs to $\phi$ and vice versa, releasing the model from the burden of distinguishing between two cities.
A previous study shows that this approach is effective in translating embeddings of one language to another in an unsupervised manner \cite{wada2018unsupervised}.
We refer to this translation method as ``\textbf{\texttt{Joint-MobLSTM}}''.

\subsubsection{Procrustes Transformation Approach}
The second approach applies the Procrustes transformation method, which is originally used in the supervised problem setting. 
Given place representations $\textbf{X}_{\psi}$ and $\textbf{X}_{\phi}$, and a dictionary of pairs of places which are ranked by their popularity (indicated by the superscript $(i)$), orthogonal Procrustes is applied to align them together into a common vector space by optimizing the following function:
\begin{equation}
    R^* = \argmin_{R^T R = I} \sum_{i=1}^N \norm{R X_{\phi}^{(i)} - X_{\psi}^{(i)}}_F
\end{equation}
where $\textbf{R}\in \mathbb{R}^{d \times d}$, and $\norm{\cdot}_F$ is the Frobenius norm.
The solution gives the best rotational alignment of the two vector spaces.
Although our original problem is in the unsupervised setting, we generate synthetic representation pairs by pairing up the top $N$ visited places from both cities. 
We use $N=500$ as the default number of place pairs, however we test its effect on translation performance in Section 4.4.4. 
We refer to this translation method as ``\textbf{\texttt{MobLSTM-P}}''.

\subsubsection{Adversarial Training Approach} \label{sec:mapping_of_place_repr}
The Procrustes method requires a dictionary of pairs of places from the two cities that are expected to be mapped closely together, however, a fully unsupervised approach is shown to work better in some settings \cite{conneau2017word,artetxe2017unsupervised}. 
The third approach uses adversarial training to learn the transition matrix $R$, which is then used for translating the representations learned via \texttt{MobLSTM} between the two cities.
A similar approach as Conneau et al. \cite{conneau2017word} is taken here, where a model is trained the discriminate between representations randomly selected from $R \textbf{X}_\phi$ and $\textbf{X}_\psi$.
$R$ is then trained to prevent the discriminator from making accurate predictions \cite{ganin2016domain}. 
The standard training procedure of deep adversarial networks is used for train the adversarial model \cite{goodfellow2014generative}. 
We refer to this translation method as ``\textbf{\texttt{MobLSTM-Adv}}''.



\section{Experimental Validation}

In our experiments, we define the sizes of the places as $r = 1000m \times 1000m$ grid cells. 
Through the qualitative analysis of the place representations in Section 4.3.3, we confirm that this spatial scale is granular enough to be able to identify specific places.
Moreover, the evacuation shelter analysis in Section 5 shows that the scale is informative enough to assist disaster relief officers in practice.  
We generated and translated representations of places (grid cells) instead of specific place of interests (POIs) that are specified in maps, because there are cases where places with no particular POI could have significant meanings to the people.

\subsection{Data}
\subsubsection{Mobile Phone Data}
Yahoo Japan Corporation\footnote{\url{https://about.yahoo.co.jp/info/en/company/}} collects location information of mobile phone app users in order to send relevant notifications and information to the users. 
The users in this study have accepted to provide their location information. 
The data are anonymized so that individuals cannot be specified, and personal information such as gender, age and occupation are unknown. 
Each GPS record consists of a user's unique ID (random character string), timestamp, longitude, and latitude. 
The data acquisition frequency of GPS locations changes according to the movement speed of the user to minimize the burden on the user's smartphone battery. 
If it is determined that the user is staying in a certain place for a long time, data is acquired at a relatively low frequency, and if it is determined that the user is moving, the data is acquired more frequently. 
The data has a sample rate of approximately 2\% of the population, and past studies suggest that this sample rate is enough to grasp the macroscopic urban dynamics \cite{yabe2016framework,nishi2014hourly}. 
Table \ref{tab:stats} shows the statistics of the dataset collected for two cities (Kumamoto and Okayama), which are the cities that we focus on in this study.  
There are around 100,000 unique active users from both areas, and their location data were analyzed to extract their home locations and staypoint locations using methods in Section 3.1. 

\begin{table}[t]
\centering
  \caption{Data statistics for the two cities}
  \label{tab:stats}
  \begin{tabular}{ccc}
    \toprule
    & \begin{tabular}{c} Kumamoto \end{tabular} & \begin{tabular}{c} Okayama \end{tabular} \\
    \cmidrule{2-3}
    \# Users & 94,053 & 119,349 \\
    \# GPS staypoints & 2,832,329 & 2,382,861  \\
    Data period & 2016/2/1$\sim$2/29 & 2018/6/1$\sim$6/30\\
    \# places $n_c$ & 2565 & 2163 \\
  \bottomrule
\end{tabular}
\end{table}

\subsubsection{Landuse Data}
To validate whether the generated and translated place representations correctly reflect the functionality of the places, we use the Urban Area Land Use Mesh Data\footnote{\url{http://nlftp.mlit.go.jp/ksj/gml/datalist/KsjTmplt-L03-b-u.html}} in the National Land Numerical Information Database\footnote{\url{http://nlftp.mlit.go.jp/ksj/}} provided by the Ministry of Infrastructure, Land, and Transport and Tourism of Japan. 
The dataset divides all urban areas of the entire country into $100m \times 100m$ grid cells, and assigns one category to each grid cell out of 17 options. 
The 17 options include farmland, residential area, business district, parks, forests, factories, public facilities, water body, open spaces, roads, railways, golf courses, etc. 
We aggregate these data into our spatial scale ($1000m \times 1000m$), thus for each place, we have a 17 dimensional vector where each element shows how many pixels of a specific land type exists in that place. 

\subsection{Experiment Settings}

\subsubsection{Model Hyperparameter Settings}
To conduct the representation learning of places described in Section 3.2, we setup the model and input data as follows.
The model consists of the embedding layer, LSTM RNN block, readout layer, and the output layer.
While the main input of the model is a sequence of staypoints representing a user's movement, we added two supplementary values, which are the timestamp of when the user had entered that place and the duration time of the stay, to incorporate time-dependency of the users' behavior.
The embeddings of staypoints were set to 96-dimensional vectors. 
The timestamp and stay duration were converted to 8-dimensional and 4-dimensional vectors respectively, and the three vectors at each step were concatenated into a 108-dimensional vector.
The LSTM RNN block scanning over the embedding sequence consists of two layers of the size 128, and the hidden vectors of both layers were fed into the readout layer of the size 96, which were then read by the output layer producing the probability distribution over staypoints for the next place prediction.
The parameter matrix of the staypoint embedding was reused as the output layer's matrix to reduce the total number of parameters and make the training data usage more efficient.
We applied dropout with the keep probability 0.7 to three points of the model: the embedding layer, readout layer, and output layer.
We continued the training for 20 epochs, evaluated performance on the validation data at the end of each epoch, and used the embedding matrix of the best model for subsequent processing.

\subsubsection{Comparative Methods} \label{compare}
We first assess the quality of place representations generated by \textbf{\texttt{MobLSTM}} and \textbf{\texttt{Joint-MobLSTM}} in Section \ref{sec:intra}.
Then, after clarifying that the generated representations accurately embed the functions of places in each city individually, we validate the performances of translation models \textbf{\texttt{MobLSTM}}, \textbf{\texttt{Joint-MobLSTM}}, \textbf{\texttt{MobLSTM-Adv}}, and \textbf{\texttt{MobLSTM-P}} in Section \ref{sec:inter}.


\subsubsection{Evaluation Metrics} Two metrics are used to evaluate the performances of the translation methods. 
Given two sets of places, we measure the average mutual norm distance and average mutual cosine similarity of the place representations of those places. 
\\
\textbf{\texttt{Average Mutual Norm Distance} (AMND).} Given 2 place representations $x_i$,$x_j \in \mathbb{R}^d$, norm distance is defined as $\text{Norm-Dist}(i,j) = \norm{x_i-x_j}$. 
The average mutual norm distance between sets of places $I$ and $J$ is defined by the following:
\begin{equation}
    s_{nd}(I,J) = \frac{1}{Z(I,J)} \sum_{i\in I} \sum_{j\in J; j \neq i} \text{Norm-Dist}(i,j)
\end{equation}
where $Z(I,J)$ is the number of unique combinations between the places in sets $I$ and $J$. 
\\
\textbf{\texttt{Average Mutual Cosine Similarity} (AMCS).}
Cosine similarity is a commonly used metric to measure the similarity between 2 place representations $x_i$,$x_j \in \mathbb{R}^d$, and is defined as:
\begin{equation}
    \text{cos-sim}(i,j) = \frac{x_i \cdot x_j}{\norm{x_i} \norm{x_j}}
\end{equation}
We use the average mutual cosine similarity to measure the similarity between representations of two sets of places.
Similar to the average mutual norm distance, average mutual cosine similarity between sets of places $I$ and $J$ is defined by the following:
\begin{equation}
    s_{cs}(I,J) = \frac{1}{Z(I,J)} \sum_{i\in I} \sum_{j\in J; j \neq i} \text{cos-sim}(i,j)
\end{equation}

\begin{figure}[t]
  \centering
  \includegraphics[width=0.6\columnwidth]{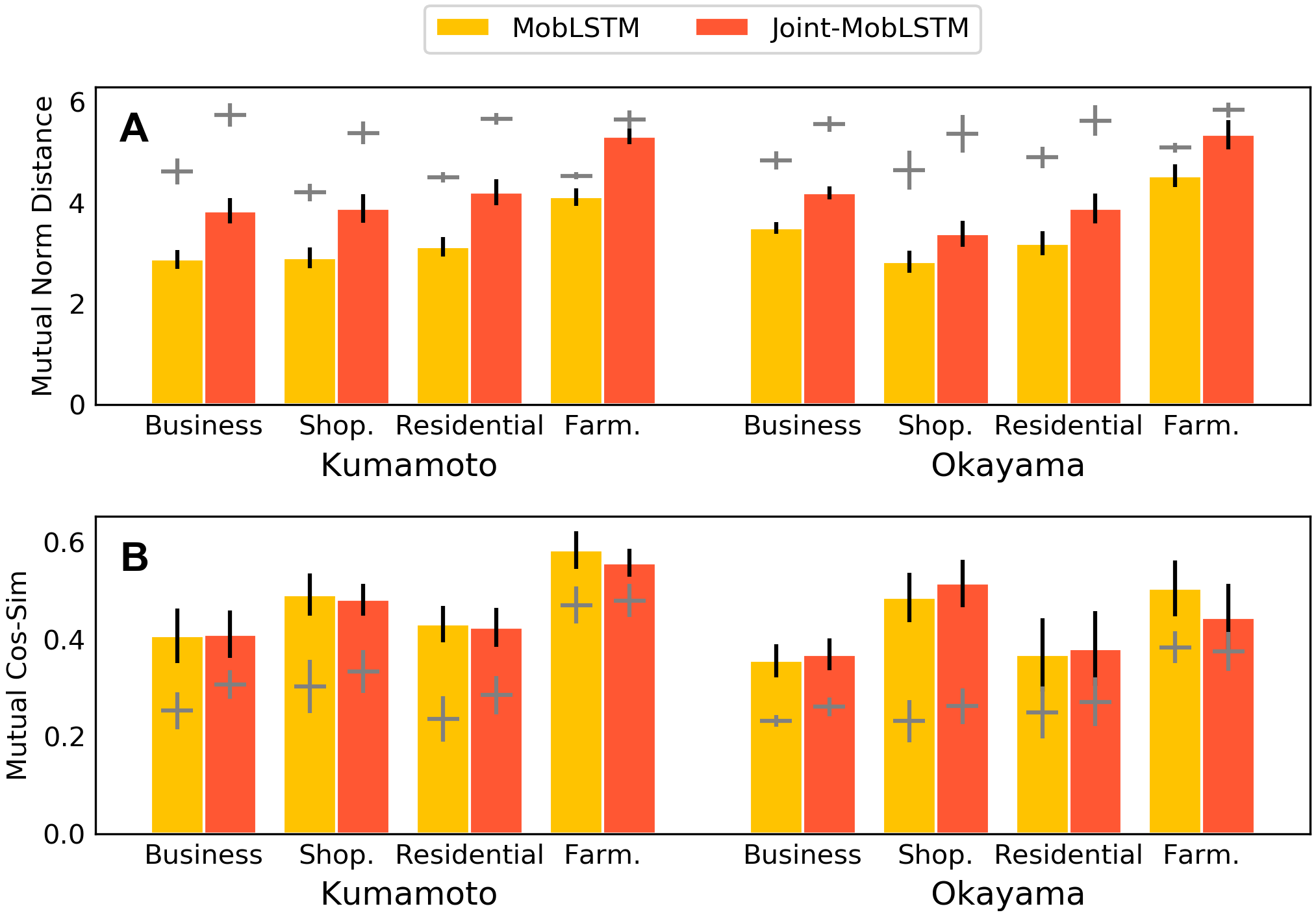}
  \caption{
  Intra-city validation results of business, shopping, residential and farmland areas in Kumamoto and Okayama with (A) AMND and (B) AMCS metrics.
  For both models, generated representations of all landuse types showed statistically significant intra-city similarity.
  }
  \label{intra}
\end{figure}

In both experiments (Sections 4.3.1 and 4.3.2), we show whether the results are statistically significant by comparing the performance metrics to random pairs of places generated by the same model.
For example, to assess the quality of place representations of \texttt{MobLSTM} for business places in Kumamoto in Section 4.3.1, we compare the AMND between representations of pairs of business places in Kumamoto generated by \texttt{MobLSTM}, against the AMND between representations of pairs of business places and randomly selected non-business places in Kumamoto generated by \texttt{MobLSTM}. 
Similarly, for example, to validate the translation performance of \texttt{MobLSTM-P} for business places in Section 4.3.2, we compare the AMND between representations of pairs of business places in Kumamoto and Okayama translated by \texttt{MobLSTM-P}, against the AMND between representations of pairs of business places in Kumamoto and randomly selected non-business places in Okayama translated by \texttt{MobLSTM-P}. 
Similarity between random pairs are shown in Figures \ref{intra} and \ref{interresults} as gray crosses (vertical line indicate error bars).



\subsection{Results}


\subsubsection{Intra-city Validation of Place Representations} \label{sec:intra}
To validate whether the place representations were correctly generated (i.e. representations of places with the same functionality are mapped closely), we measured the AMND and AMCS between places with the same landuse labels (business districts, shopping malls, residential areas, and farmland areas). 
We refer to this validation scheme as ``intra-city validation''. 
Figure \ref{intra} shows the validation results for both cities. 
Note that for norm distance (A), lower is better, and for cosine similarity (B), higher is better. 
All error bars (vertical lines) show the standard deviation of the results of 10 iterations.
Results show that both models were able to generate accurate representations, and embedded places with same landuse labels closer to eachother than randomly selected pairs of places. 
\texttt{MobLSTM} and \texttt{Joint-MobLSTM} had comparable performances for generating place representations, however we see that \texttt{MobLSTM} had slightly better performances (lower norm distances and higher cosine similarity) for many of the landuse types.
This result agrees with our intuition, because \texttt{MobLSTM} is able to allocate more dimensions in the parameter space to encode information related to places in their own city, whereas \texttt{Joint-MobLSTM} shares the parameter space across different cities, having less dimensions to encode the representations for each city. 

\begin{figure}[!t]
  \centering
  \includegraphics[width=0.6\columnwidth]{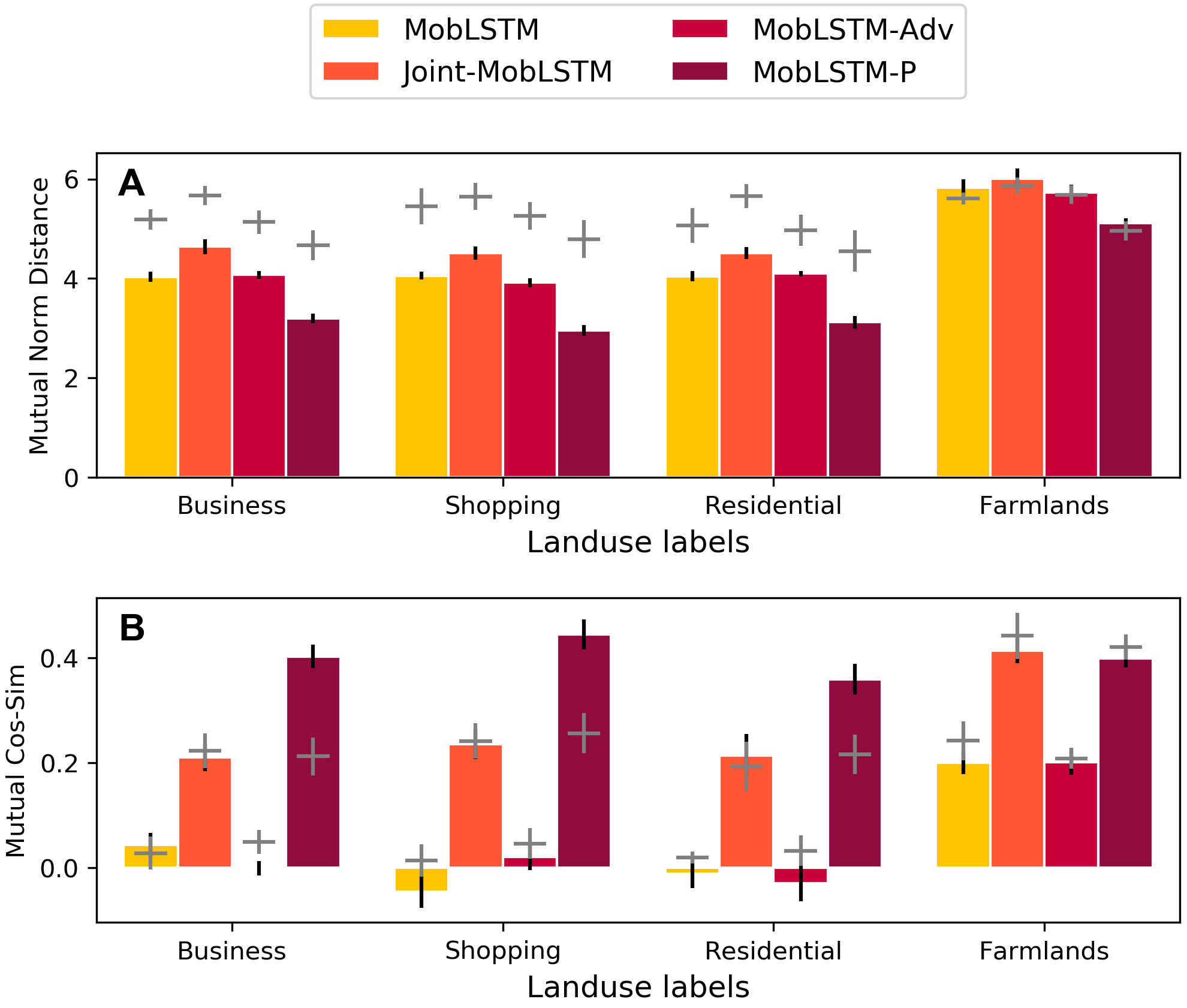}
  \caption{
  Inter-city translation results from Kumamoto to Okayama of business, shopping, residential and farmland areas with (A) AMND and (B) AMCS metrics. 
  \texttt{MobLSTM-P} is able to translate place representations across Kumamoto and Okayama for business, shopping, and residential areas, but not for farmland areas where little human mobility patterns are observed. 
  }
  \label{interresults}
\end{figure}

\begin{figure*}[!t]
  \centering
  \subfloat{\includegraphics[width=0.5\textwidth]{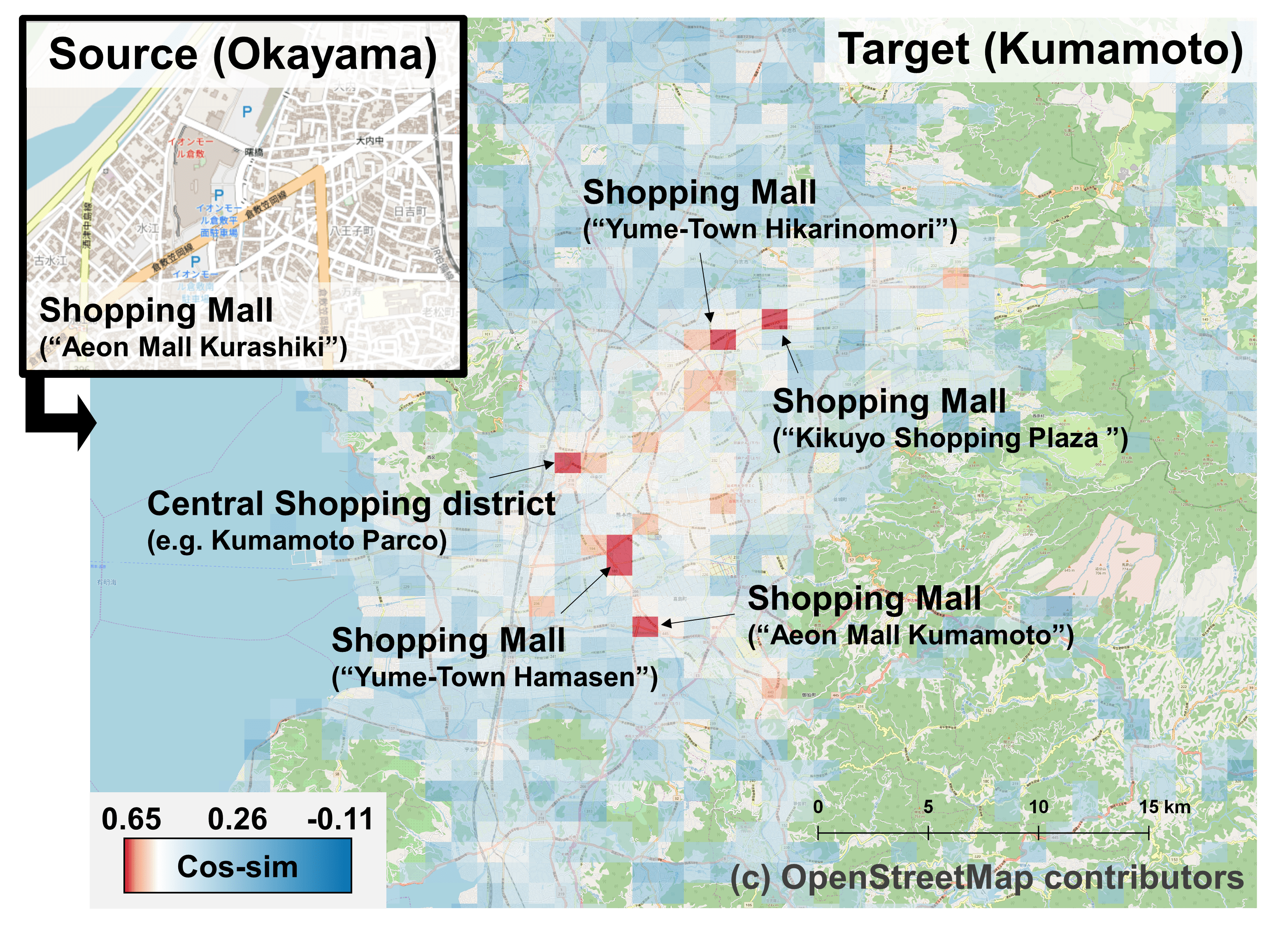}}
  \subfloat{\includegraphics[width=0.5\textwidth]{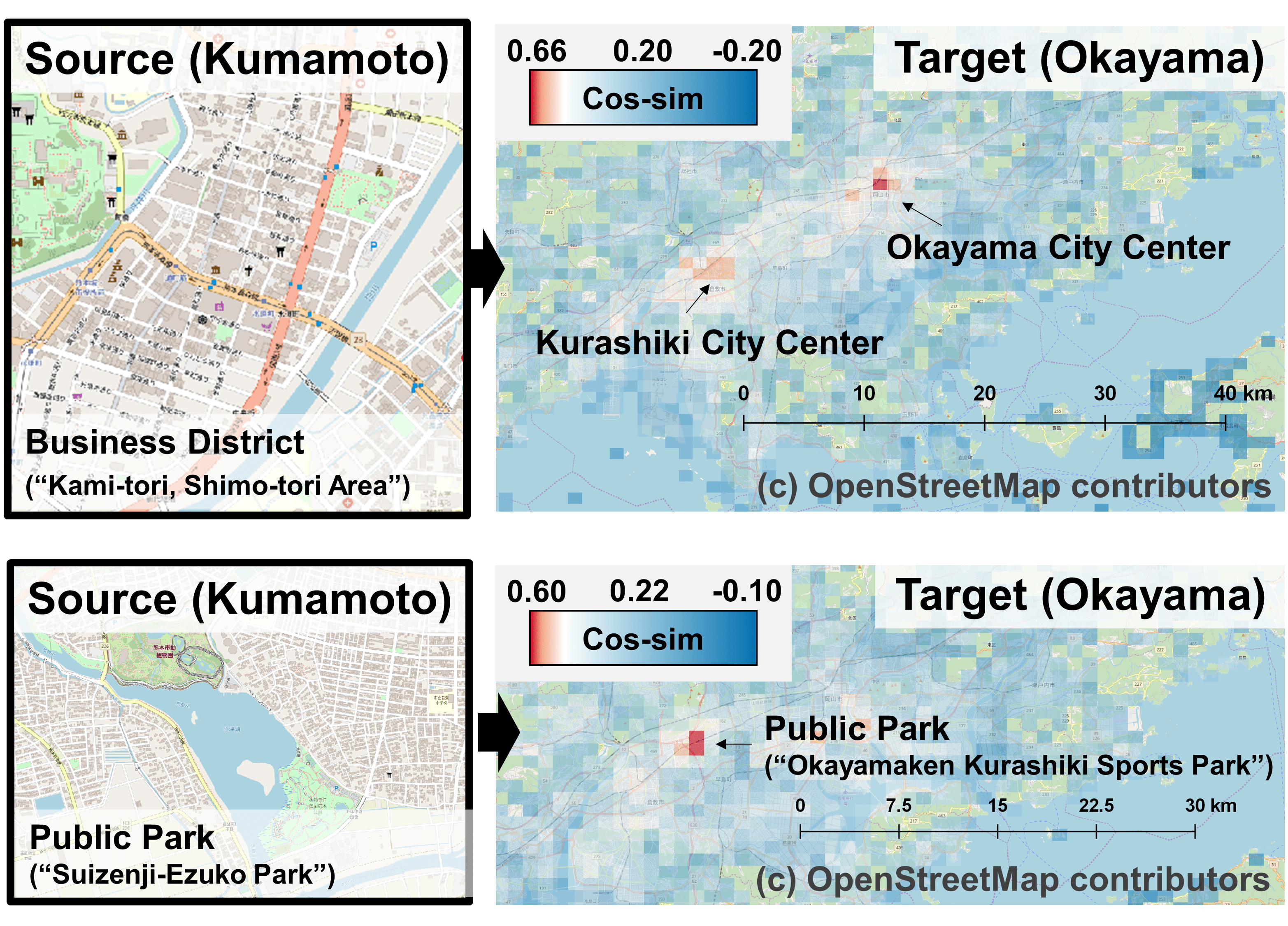}}
  \caption{
  Qualitative analysis and inspection of the translated place representations across cities using \texttt{MobLSTM-P}. 
  The three panels show that for both directions (Kumamoto $\rightarrow$ Okayama and vice-versa), places such as shopping malls, business districts, and public parks were successfully translated so that places with similar functions from different cities were mapped closely together in the common vector space.  
  }
  \label{aeonkurashiki}
\end{figure*}

\subsubsection{Inter-city Translation of Place Representations} \label{sec:inter}
To quantitatively validate the performance of translating place representations across cities, we measured the AMND and AMCS between places with same landuse types across different cities (e.g. similarity between representations of places with shopping malls in Kumamoto and representations of places with shopping malls in Okayama). 
We refer to this validation scheme as ``inter-city translation''. 
Figure \ref{interresults} shows the translation accuracy of all tested methods. 
Out of the four models, \texttt{MobLSTM} has had no translation operation, and in all landuse types the AMCS performance is worse than random, which confirms the negative example illustrated in Figure \ref{problem} does occur, and that a translation operation is indeed needed to compare place representations from different cities. 

The rest of the models compare the performances of different translation methods. 
For business, shopping, and residential areas, the AMND seems to be lower than random for all methods, which implies that all of these types of places are clustered together in the vector space, away from the farmland areas. 
The AMCS metric allows us to measure more specific  differences in the place representations, by normalizing the vectors by their lengths.
AMCS results show that \texttt{MobLSTM-P} is able to translate representations of business, shopping and residential places successfully (statistically significantly).
Even though \texttt{Joint-MobLSTM} and \texttt{MobLSTM-Adv} approaches are shown to succeed in language translation tasks, they fail to do so in place translation tasks.
The failure of \texttt{Joint-MobLSTM} implies that the model was complex enough to completely distinguish places between Kumamoto and Okayama, and to embed them separately in the common vector space, contrary to our intuition. 
For \texttt{Joint-MobLSTM} to perform better, further searching for the appropriate model architecture may be effective, however is not cost effective considering the vast model space. 
\texttt{MobLSTM-Adv} failing to align the two vector spaces indicates that the probability distribution of the place representations of two cities are completely different, in contrary to word vector spaces. 
Even with \texttt{MobLSTM-P}, representations of farmlands were not successfully translated across cities. 
This is because we are not using many farmland areas as anchor points in our dictionary for solving the optimization problem (Section 3.3.2), due to the lack of observed human mobility patterns in such areas.
Overall, results in Figure \ref{interresults} confirm that \texttt{MobLSTM-P} is successful in translating representations of places visited by people (business, shopping, and residential) across cities. 


\subsubsection{Qualitative Inspection of Translated Representations}
In addition to the quantitative evaluation, we inspected whether the place representations translated by \texttt{MobLSTM-P} were actually mapped close to similar places in the target city. 
Figure \ref{aeonkurashiki} shows successful cases where the place of the source city is mapped closely with similar locations in the target city. 
In each of the three panels, the original place in the source city is shown in the left black box (e.g. Shopping Mall ``Aeon Mall Kurashiki'' of Okayama), and the cosine-similarity values between the representations of all places in the target city and the translated representation of the original place is shown in color on the map. 
Red and blue colors show high and low similarity, respectively.
The color bar is adjusted so that only the places with top 5 percentile cosine similarity are shown in red. 
For places with high similarity (red places), POIs within each place are annotated on the map. 

The left panel shows how a shopping mall in Okayama, when translated to the Kumamoto vector space, becomes mapped close to major shopping malls in Kumamoto, including the central shopping district, Aeon Mall Kumamoto, and several other shopping facilities. 
The two panels on the right side show how translation of places in the opposite direction (Kumamoto $\rightarrow$ Okayama) also produced intuitive and accurate results.
The top right panel shows that the business districts of Kumamoto (the Kami-tori and Shimo-tori area), when translated to the Okayama vector space, were similar to the two major city center districts (Okayama city center and Kurashiki city center), implying that urban functionality can also be translated successfully, reinforcing our results in Figure \ref{interresults}.
The bottom right panel shows an instance where even a major public park in Kumamoto (``Suizenji-Ezuko Park'') was successfully translated so that it became mapped close to a major park in Okayama (``Okayamaken Kurashiki Sports Park''). 
Although public parks were not included in our quantitative evaluation, this result implies that \texttt{MobLSTM-P} can translate more specific places of interest to other cities.
Overall, Figure \ref{aeonkurashiki} shows promising results that \texttt{MobLSTM-P} successfully maps similar places together onto the common vector space.

\begin{figure}[t]
  \centering
  \includegraphics[width=0.6\columnwidth]{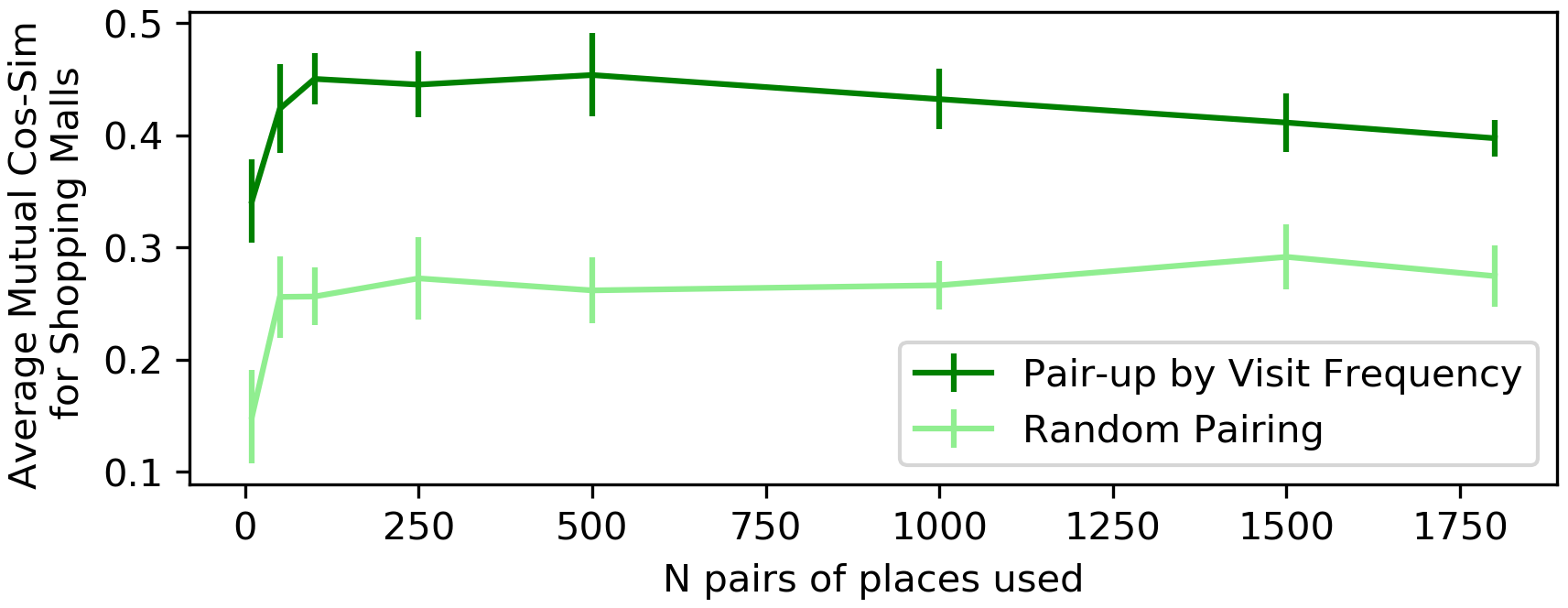}
  \caption{
    Translation performance of shopping malls from Kumamoto to Okayama using \texttt{MobLSTM-P}, using different number of anchor places for translation (x-axis), chosen by 2 different criteria (most frequently visited or random). 
  }
  \label{sensitivity}
\end{figure}

\subsubsection{Sensitivity to Number of Place Pairs}
The \texttt{MobLSTM-P} model requires a synthetic dictionary (a dataset with pairs of places) used to solve the optimization task. 
In the previous sections, we used $N=500$ as the number of anchor places (i.e. pairs of places used for the optimization task). 
Here we show a sensitivity analysis of this parameter, by looking at the average mutual cosine similarity of shopping places in Kumamoto and Okayama using \texttt{MobLSTM-P} with different values of $N$.
We observe from Figure \ref{sensitivity} that the performance initially increases as we increase the number of pairs. 
However we see a plateau in performance after $N=100$, and a decrease after $N=500$, implying that choosing too many places as anchors adds too much irrelevant information for choosing the optimal rotation. 
Even though our method beats random pairing (light green color) for all $N$ values, this result indicates that selecting the appropriate number of anchor places is important for the performance of our method.
Further investigation using data from more cities is needed to determine whether there is a universal rule in determining the appropriate parameter value for $N$.

\section{Discussions}
In this paper, we proposed and tested methods to translate place representations across cities. 
Experimental results using real world mobility data from two cities in Japan clarified that we are indeed able to translate representations of places across cities accurately using \texttt{MobLSTM-P}, which finds the best rotational alignment between vector spaces using anchor places based on visit frequencies through optimization.
We clarified both quantitatively and qualitatively that places from different cities with similar landuse types became mapped closely in the common vector space after translation. 
Moreover, although the task of translating place representations across cities is analogous to word translation across languages, we observed several differences in the problem setting through failures of methods that were successful in the language translation domain, namely the joint learning (\texttt{Joint-MobLSTM}) and adversarial learning (\texttt{MobLSTM-Adv}) approaches. 
In addition to evaluating the translation performances, we showcased a case study of an important urban challenge that may be better approached using our inter-city translation method, which was to use the representations of evacuation shelter locations from a disaster in the past to predict evacuation shelters in a future disaster in another city. 

Now, we discuss future research opportunities that this study enables. 
The first direction of research is on improving the accuracy of the translation task. 
In this study, we tested several methods that extended the state of the art methods for unsupervised machine language translation developed in the natural language processing field.
However, we believe that we are able to improve the accuracy by further integrating characteristics specific to geographical locations, that are different from words and sentences.
For example, words have stronger interchangeability characteristics, since words can often be interchanged with very little or even no cost at all (\textit{I have a \textbf{cat}} $\leftrightarrow$ \textit{I have a \textbf{dog}}).
However, that is much less likely in sequence of places and it is much rare to have two or more places with exact interchangeability in the routines for human beings. 
Integrating insights from the human behavioral sciences into building the representation learning and translation model would be an interesting topic for future studies. 

The second direction of research is to increase the diversity of cities for testing.
Although the finding that we are able to translate place representations across different cities (Kumamoto and Okayama) was insightful and promising, we are motivated in further investigating whether this method works between a more diverse set of cities, such as Tokyo, Japan and Indianapolis, USA, where various aspects (e.g. social norms, peoples' mobility patterns, city structures) are more different than between Kumamoto and Okayama. 
Should the method fail in such diverse pairs of cities, developing new models that consider exogenous contexts via fusion with other data sources would be an important and interesting problem. 
We hope to utilize a larger mobility dataset to investigate this topic in future studies. 

We would also like to look into potential problems where we can apply this technique.
Selection of appropriate locations to open new stores has been a popular problem in urban planning \cite{xu2016}. 
Testing whether translating successful/unsuccessful locations across cities could predict success/failure of new stores, is of future research interest.

\section{Related Works}
\subsection{Place Representation Learning}
Learning the representations of places have been a popular research topic in the field of urban computing \cite{zheng2014urban}, often as a subproblem for larger tasks such as POI recommendation \cite{chang2018content} and site selection problems \cite{xu2016}.
Recent developments in the natural language processing field on representation learning, such as \texttt{word2vec} \cite{mikolov2013distributed}, has inspired many studies on place representation learning.
Models such as \texttt{SkipGram} and \texttt{POI2vec} have applied ideas similar to \texttt{word2vec} on social media check-in data, where sequences of POIs are treated as sentences in the \texttt{word2vec} model \cite{liu2016exploring,feng2017poi2vec}. 
\texttt{CAPE} also used Instagram check-in data, and also uses text data for embedding the POIs \cite{chang2018content}. 
\texttt{Geo-Teaser} uses the users' check-in data and also considers the geographical proximity of POIs as well when generating representations \cite{zhao2017geo}. 
\texttt{Place2vec} does not use the user check-in sequences, but instead uses the physical proximity between POIs and the number of visit counts to increase training data so that such pairs of POIs would have similar embeddings \cite{yan2017itdl}.

With the availability of large scale mobility data such as taxi GPS data and mobile phone data, more recent studies such as \texttt{DeepMove} have extended the aforementioned methods to spatio-temporal richer datasets \cite{zhou2018deepmove}.
Wang et al. \cite{wang2017region} have extended the idea to learn the representations from mobility flow modeled as flow graphs, to include multi-hop transitions in their embeddings.
Using the New York Taxi GPS dataset, \texttt{ZE-Mob} proposes a origin-destination coupled embedding model, where the assumptions are that origin and destinations of the same trip should have similar representations, and that trips taken in similar timeframes are similar \cite{yao2018representing}.
Since the main focus of this paper is to show that we can \textit{translate} the produced representations across cities, we do not replicate and compare the performances of all of the models listed here. 
Instead, we apply a method similar to \texttt{ZE-Mob} in this study to produce place representations from mobility data. 

\subsection{Unsupervised Machine Translation}
Machine translation, especially in the unsupervised setting, has become a popular topic in the representation learning literature, due to the difficulty of collecting large scale cross-lingual training data \cite{artetxe2017unsupervised,artetxe2018unsupervised}. 
This unsupervised setting applies to our problem setting, since we are not given any training data in the form of location pairs from different cities to learn what places should be closely embedded.
Studies take different approaches to unsupervised word translation tasks; Zhang et al. \cite{zhang2017adversarial} uses an adversarial training approach, Conneau et al. \cite{conneau2017word} learns a linear transformation matrix to map words in one language to another,  Artetxe et al. \cite{artetxe2018robust} applies better initial estimates using probability density functions of distances to other word embeddings, and Wada et al. \cite{wada2018unsupervised} apply a shared \texttt{LSTM} model to jointly embed two languages. 
Cross-comparative studies of these methods have shown that the \texttt{LSTM} models works best under low resources (training data). 
Hamilton et al. \cite{hamilton2016diachronic} apply a similar idea with Conneau et al. \cite{conneau2017word} to align word embeddings computed from corpus from different years, to understand the transition of the semantics of words over the years.

In the context of urban mobility, Pang et al. \cite{pang2018replicating} applied a reinforcement learning approach in replicating the usual urban dynamics.
The study closest to ours is \texttt{CityCoupling}, which attempts to map human mobility trajectories from one city to another \cite{fancoupling}. 
Although the framework replicates the mobility patterns on New Years Day in another city well, the model does not focus on translating the functions of different locations. 
Moreover, the simulation results of the human mobility trajectories if the Great East Japan Earthquake happened in Osaka, lacks validation.
In this paper, we apply and extend the methods developed for unsupervised machine translation to translate place representations across different cities and test their performances using real world data. 

\section{Conclusion}
Despite the popularity in understanding urban functions through place representations generated from mobility patterns, no works so far have attempted to translate the representations across different cities. 
To bridge this gap, we tested methods inspired by machine language translation to translate place representations across cities. 
Experimental results show that our methods were able to translate place representations, so that functionally similar places from different cities were mapped closely together in the common vector space. 
The experimental results as well as the case study on disaster shelter prediction shows that this avenue of research may offer a novel and effective approach on solving various urban challenges via knowledge sharing across different cities.

\bibliographystyle{unsrt}  
\bibliography{references}  

\end{document}